\documentclass[conference]{IEEEtran}
\IEEEoverridecommandlockouts
\usepackage{amsmath,amssymb,amsfonts}
\usepackage{graphicx}
\usepackage{hyperref}
\begin{document}

\title{Using remotely sensed data for air pollution assessment}

\author{\IEEEauthorblockN{Teresa Bernardino}
\IEEEauthorblockA{\small INESC-ID Lisboa\\
\\
Instituto Superior Técnico \\
Universidade de Lisboa, Portugal\\
{\small 	teresabernardino@tecnico.ulisboa.pt}
}
\and
\IEEEauthorblockN{Maria Alexandra Oliveira}
\IEEEauthorblockA{\small Centre for Ecology, Evolution and Environmental Changes \\
CHANGE - Institute for Global Change and Sustainability, \\
Faculdade de Ciências\\
Universidade de Lisboa, Portugal \\
{\small maoliveira@ciencias.ulisboa.pt}
}

\and
\IEEEauthorblockN{João Nuno Silva}
\IEEEauthorblockA{\small INESC-ID Lisboa\\
 \\
Instituto Superior Técnico \\
Universidade de Lisboa, Portugal\\
{\small joao.n.silva@inesc-id.pt}
}
}

\maketitle

\begin{abstract}

Air pollution constitutes a global problem of paramount importance that affects not only human health, but also the environment. The existence of spatial and temporal data regarding the concentrations of pollutants is crucial for performing air pollution studies and monitor emissions. However, although observation data presents great temporal coverage, the number of stations is very limited and they are usually built in more populated areas.

The main objective of this work is to create models capable of inferring pollutant concentrations in locations where no observation data exists. A machine learning model, more specifically the random forest model, was developed for predicting concentrations in the Iberian Peninsula in 2019 for five selected pollutants: $NO_2$, $O_3$ $SO_2$, $PM10$, and $PM2.5$. Model features include satellite measurements, meteorological variables, land use classification, temporal variables (month, day of year), and spatial variables (latitude, longitude, altitude).

The models were evaluated using various methods, including station 10-fold cross-validation, in which in each fold observations from 10\% of the stations are used as testing data and the rest as training data. The $R^2$, RMSE and mean bias were determined for each model. The $NO_2$ and $O_3$ models presented good values of $R^2$, 0.5524 and 0.7462, respectively. However, the $SO_2$, $PM10$, and $PM2.5$ models performed very poorly in this regard, with $R^2$ values of -0.0231, 0.3722, and 0.3303, respectively. All models slightly overestimated the ground concentrations, except the $O_3$ model. All models presented acceptable cross-validation RMSE, except the $O_3$ and $PM10$ models where the mean value was a little higher (12.5934 $\mu g/m^3$ and 10.4737 $\mu g/m^3$, respectively).

\end{abstract}


\section{introduction}

     Air pollution is a serious problem that affects not only human health, but also the environment, on a global scale. Many studies have been conducted regarding air pollution, which focus on researching the adverse effects of pollutant exposure on human health. These studies show that exposure to various air pollutants leads to increased mortality due to lung cancer, pulmonary disease, cardiovascular disease, and acute respiratory infections \cite{dockery1993association, WHO_air_quality_health, EEA_air_quality_report}.  

    In addition to the serious effects on human health, air pollution also contributes to the degradation of the environment, causing long lasting consequences on the ecosystems and biodiversity \cite{EEA_air_quality_report}. Pollutants in the atmosphere are transported to the surface, either through wet or dry means, in a natural process denominated atmospheric deposition. Atmospheric deposition can result in acidification, eutrophication and accumulation of toxic substances, which are very damaging to the ecosystems \cite{EEA_air_quality_report}.

    Given the grievous effects of air pollution in human health and the environment, tackling this problem became of paramount importance. Many organizations have developed several approaches to limit the emissions of air pollutants and increase air quality, such as creating guidelines and policies to mitigate the effects of air pollution on both human and ecosystem's health. 

    The existence of spatial and temporal data regarding the concentration and deposition of pollutants, as well as tools capable of processing the data, is vital, not only to various countries, but also to researchers to predict how changes in air pollution will affect both human health and ecosystems, and how to mitigate these effects. 

    The three main pollutant data sources include: observations from ground monitoring stations that collect pollutant concentration and deposition with varying frequency (from hourly to monthly); chemical transport models that simulate emissions, chemical exchanges in the atmosphere and interaction with the climate (e.g., precipitation, wind) and land use, with time steps ranging from hourly to annual; and remote sensing that capture reflectance's at specific wavelengths of the electromagnetic spectra reflecting pollutant total/tropospheric columns at the instant of the satellite sweep.

    Monitoring stations provide reliable data on the concentration of pollutants with great temporal resolution, since measurements are usually acquired every hour. However, the spatial distribution of air quality stations is lacking due to the associated high costs of building, operating and maintaining each station. Therefore, the number of stations is very limited and they are usually built in more densely populated areas, leaving a large portion of the territory without any pollutant concentration measurements \cite{european2011application}.

    Chemical transport models are used to complement the data acquired in monitoring stations, since they provide data with large temporal and spatial coverage. Nonetheless, chemical transport models also present some limitations. One limitation is that these models require extensive input data, such as meteorological and emissions, which can be unavailable or unreliable. In fact, climate variables and emissions are themselves obtained using models. Another limitation is the fact that the complex physical and chemical processes are simulated using simpler numerical and empirical equations, and so, model results need to be evaluated for each region before being used, which can be a tall order given the lack of ground measurements \cite{european2011application}.

    In order to solve some of the limitations of the ground based measurements from monitoring stations and chemical transport model results, remote sensing data is also integrated to complement these two, since it provides global coverage with good spatial resolution. However, the temporal resolution of remote sensing data is lower than ground measurements, since satellites acquire data daily, whereas monitoring stations acquire data every hour. Also, sometimes there are no data retrievals due to cloud cover \cite{veefkind2007applicability}. Another disadvantage is that remote sensing retrieves the total column of each substance, and so, to determine surface-level concentrations, ground measurements and remote sensing data need to be related using statistical models or machine learning algorithms \cite{veefkind2007applicability}.

    Many different groups of people from various backgrounds need access to air pollution data such as epidemiologists, policy makers, researchers, etc. However, this data is not always easily accessible to most people, given the numerous data sources and formats available, complex download, lack of tools to appropriately process and aggregate the data, or the complexity associated with learning each new different tool/dataset. As a result, use of air pollution data for either research or governance ends up being much more time consuming and complex due to data handling, which makes it a very difficult and tiresome process.

    One of the main objectives of this work is, not only to study the application of remote sensing in the field of air pollution, but also to create models capable of inferring pollutant concentrations in locations where no observation data exists, in an attempt to create new data sets concerning air pollution. Another objective of this work is to facilitate the access to air pollution data, by creating libraries that automate the download, aggregation and processing of various data regarding air pollution.

    Firstly, various python scripts were created to process different air pollution data. For the selected air pollution data sources, the download, aggregation, cleaning and other processing of the respective data were automated, allowing ease of access to additional air pollution data. The chosen data sources include: European Air Quality Portal \cite{EAQP_webpage}, for concentration of pollutants measured in air quality stations; and Sentinel-5P Pre-Operations Data Hub \cite{esa_sentinel5p_data_hub}, for remotely sensed data. 

    Secondly, Sentinel-5P data for the years 2018 and 2019, covering the Iberian Peninsula, was downloaded and processed. The selected data products include: Ozone total column (L2\_\_O3\_\_\_\_), Nitrogen Dioxide total and tropospheric columns (L2\_\_NO2\_\_\_), Sulfur Dioxide total column (L2\_\_SO2\_\_\_), and UV Aerosol Index (L2\_\_AER\_AI). 

    Lastly, a machine learning model was created to infer pollutant concentrations in the Iberian Peninsula in locations where no observations exist. The developed model is a random forest model, that includes as features satellite measurements, meteorological variables, land use classification, and other temporal variables, such as, month, day of year, etc. Meteorological, land use and temporal variables were included not only to account for the spatial and temporal variation of pollutant concentrations, but also the effect on the creation and transport of pollutants. Wind and precipitation greatly influence the dispersion of pollutants. Different land uses include urban, industrial, and transport classifications, that identify locations where most emissions occur. Meteorological and land use data sources were analyzed to create the machine learning model, including: ERA5 dataset from the Climate Data Store \cite{era5_hourly_single_levels} for various meteorological variables; and Corine Land Cover dataset from the Copernicus Land Monitoring Service \cite{corine_land_cover_webpage} for land use classification data. The download and processing of the respective data was also automated. The developed machine learning model was evaluated temporally and spatially using three different methods: cross-validation, each fold using 10\% of the observations from 2019 as test data; using data from 2019 as train data and from 2018 as test data; and station 10-fold cross-validation using part of the stations as test data and the rest as train data, utilizing data from 2019.

\section{Related Work}

    \subsection{Air Pollution}
    
        \hspace{\parindent}
        Air pollution is defined as a contamination of the environment by any chemical, physical or biological agent that contributes to altering the natural characteristics and composition of the atmosphere, resulting in serious consequences for human health and the environment \cite{WHO_air_quality_health}. The amount of pollutant in a standard volume of a certain medium, such as air or water, is normally measured and referred as pollutant concentration. 

        Air pollutants can be classified as primary or secondary, depending on their origin. Primary pollutants are emitted into the atmosphere, whereas secondary pollutants result from chemical reactions and other processes between primary pollutants \cite{EEA_air_quality_report, eea_air_pollution_sources}. 

        Some air pollutants are extremely hazardous to human health and the environment, and, as a result, are highly discussed in the literature. These air pollutants include: particulate matter with a diameter of 10 micrometres or less ($PM10$) and 2.5 micrometres or less ($PM2.5$), ozone ($O_3$), nitrogen oxides ($NO_x$, which comprise nitrogen monoxide ($NO$) and dioxide ($NO_2$), methane ($CH_4$), carbon monoxide ($CO$), ammonia ($NH_3$), and sulphur dioxide ($SO_2$). Main sources of emissions of air pollutants include: fuel combustion, industrial processes, agricultural activities and waste treatment \cite{EEA_air_quality_report, eea_air_pollution_sources}.

    \subsection{Air pollution data sources}
    
        \hspace{\parindent}
         Air pollution data can be acquired from some different main sources, one of them being ground measurements from monitoring stations. Air quality stations can be classified as one of three types, depending on their location and the presence/absence of local emission sources, these being: background, where measurements are not influenced by daily fluctuations originated from industrial or urban areas; industrial, where measurements are essentially influenced by nearby industrial emissions; and traffic, where measurements are significantly influenced by nearby traffic emissions \cite{geiger2013assessment}.

        Air quality stations directly and continuously measure the concentration of major pollutants, acquiring hourly measurements, and, as a result, providing great temporal coverage. Monitoring stations were the first approach to monitor air pollution, and the data acquired is still considered the one that more accurately portrays reality \cite{european2011application}. Nevertheless, the number of monitoring stations is very limited due to the associated high costs of constructing, operating and maintaining a station. Since most stations are built in more densely populated areas, a large portion of the territory doesn't have any ground measurements, and so, monitoring stations provide air quality data with poor spatial coverage \cite{european2011application}.

        Chemical transport models simulate the atmospheric chemistry, replicating the physical and chemical processes air pollutants are subject to when in the atmosphere. These models offer good temporal and spatial coverage, and so, are used to complement ground measurements from air quality stations, since they can provide air pollution data for places where no concentration measurements exist. Models also enable the development of guidelines and policies for emission reduction scenarios, since they can predict pollutant concentrations when analysing emission changes \cite{european2011application}.
  
        Nonetheless, chemical transport models require considerable input data, such as meteorological conditions, land cover, and emissions data, which can be unreliable or unavailable. Additionally, the complex physical and chemical processes are simulated using simpler numerical and empirical equations, further increasing the uncertainty of the results \cite{european2011application}. 
        
    \subsection{Remote sensing}
    
        \hspace{\parindent}
        Remote sensing is the process of acquiring information regarding the characteristics of an object or an area, at a distance, by analysing the emitted and reflected radiation from Earth with sensors aboard aircraft or satellites \cite{veefkind2007applicability,nasa_earthdata_remote_sensing}. 

        Remotely sensed data and its application can vary depending on the resolution of the data, which is directly related to the satellite orbit and the utilized sensor. Resolution encompasses four types: spatial, temporal, spectral and radiometric \cite{nasa_earthdata_remote_sensing}. Spatial resolution refers to the size of a pixel on the raster dataset, each pixel representing a specific area on Earth. The higher the spatial resolution, the more detail will be captured in the dataset \cite{nasa_earthdata_remote_sensing}. Temporal resolution refers to the time taken to complete an orbit and return to the same position on the globe. Spatial and temporal resolution are directly related, and there is a trade-off between the two. For instance, to achieve high temporal resolution, the orbit swath needs to be larger to cover more ground in less amount of time, however larger swath results in lower spatial resolution \cite{nasa_earthdata_remote_sensing}. Spectral resolution is the ability of a sensor to distinguish finer wavelengths. The higher the spectral resolution, more bands covering finer wavelengths there will be, and, as a result, the objects that are being observed can be more easily discriminated and detailed \cite{nasa_earthdata_remote_sensing}. Radiometric resolution indicates the number of bits that can be used to store different information regarding the energy level that was measured. A finer radiometric resolution allows to store more information and distinguish between energy levels with small differences giving more detailed information \cite{nasa_earthdata_remote_sensing}.

        Remote sensing can be applied to numerous areas of study, one of them being air quality monitoring. Different chemical compositions and particles in the atmosphere scatter and absorb light differently, and depending on the type and size of particle/atmospheric composition, the reflected light will have particular wavelengths \cite{veefkind2007applicability}. Sensors aboard satellites are capable of quantifying chemical compositions and particles that absorb, scatter and emit radiation within defined bands of the electromagnetic spectrum \cite{veefkind2007applicability}. Regarding air quality, satellites retrieve the total column of a substance, that is, the integrated concentration of the substance from ground level to the top of the atmosphere \cite{veefkind2007applicability}.

        Remotely sensed data can be used to complement ground measurements from air quality stations and data obtained from chemical transport models, since it provides global coverage with good spatial resolution, thus allowing the monitoring of locations where no observations exist \cite{veefkind2007applicability,engel2004recommendations}. Nonetheless, remotely sensed data presents some limitations. The temporal resolution is lower than ground measurements, the best resolution achievable being daily for measurements covering the whole globe, whereas monitoring stations provide hourly data. In addition, the presence of clouds largely affects the quality of satellite retrievals, as a result, data might not be available for all satellite passages \cite{christopher2010satellite}. Another limitation is that remote sensing doesn't measure directly pollutant concentrations, since remote sensing measurements correspond to the integrated concentration of the pollutant from ground level to the top of the atmosphere, which need to be transformed using ground data to determine actual ground concentrations \cite{veefkind2007applicability,christopher2010satellite}.

        Many studies have been conducted to create models capable of inferring ground level pollutant concentrations from column measurements retrieved by satellites. Many differing approaches have been implemented, such as, simple linear regression models, or more complex statistical models, which may include meteorological variables to account for the direct affect of these variables on the creation, transport and deposition of pollutants \cite{mirzaei2020estimation,qin2017estimating,lee2012use}. Given the non-linear relation between satellite measurements and ground observations, machine learning approaches, such as random forest model or neural networks, are widely used, achieving better results and also integrating meteorological, land cover and temporal variables \cite{chen2019comparison,chen2018machine}.
        
    \subsection{Machine Learning}
    
        \hspace{\parindent}
        
        Machine learning is an evolving area of study dedicated to creating algorithms and models capable of automating and optimising various tasks in areas where developing conventional algorithms to perform these tasks would be very challenging \cite{bonaccorso2017machine,mahesh2020machine}. Unlike conventional algorithms, machine learning algorithms are not explicitly programmed, learning from experience by analysing input data \cite{bonaccorso2017machine,mahesh2020machine}.
        
        Machine learning approaches have been applied to many different areas of study, one of them being air pollution, in which various models have been created to estimate pollutant concentrations in locations or temporal intervals where ground measurements are lacking or non-existent. The problem of estimating pollutant concentrations can be classified as a supervised learning problem, since input data comprises of various features, such as location data, meteorological data, land classification, etc., and ground measurement pairs, where the model learns directly from the desired output for each training example. The problem is also classified as a regression problem, since the output, the pollutant concentration, is a continuous variable. 
        
        One of the machine learning approaches that has been applied to estimate pollutant concentrations is the support vector machine (SVM) model. Chi-Man Vong \textit{et al.} \cite{vong2012short} applied the support vector machine algorithm to develop a model capable of determining short term predictions of pollutant concentrations for various pollutants, such as, particulate matter, ozone, nitrogen dioxide, and sulphur dioxide. For each pollutant, the authors experimented with multiple different kernel functions, which included: linear, polynomial, radial basis function (RBF), wavelet, and sigmoid kernels. The developed models were evaluated based on some error measures, such as, mean absolute error, root mean squared error, and relative error. The best performing models utilized the linear or the radial basis function kernels, both having similar performances. However, the models that utilized the polynomial or wavelet kernels presented very poor results, and models that utilized the sigmoid kernel presented even worse results, with much bigger error values. The authors concluded that the careful tuning and choice of kernel if of utmost importance in a support vector machine model development. The authors also attributed the lower performance to the existence of more hyperparameters, that they regarded as very difficult to optimize. Suárez Sánchez \textit{et al.} \cite{sanchez2011application} created a model based on support vector machine to study air quality at a local scale in an urban area in Spain. Similar to the previous study, the authors also concluded that tuning the hyperparameters and chosing the kernel were a crucial part in the development of the model.
        
        Another machine learning approach utilized in the field of air pollution is the artificial neural network. Jie Chen \textit{et al.} \cite{chen2019comparison} developed and compared 16 models using differing techniques, such as, linear regression, regularization, and machine learning algorithms, to predict annual average PM2.5 and nitrogen dioxide concentrations in Europe. The developed models used as predictors satellite measurements, chemical transport model results, and land use classification variables. The models were evaluated by executing 5-fold cross-validation and external validation using two different ground measurement datasets. The authors verified that after applying cross-validation, the artificial neural network model presented the lowest $R^2$ and the highest RMSE of all the tested machine learning models. The authors attributed the lower performance of the neural network model to the simple structure used, which included only one hidden layer, and primarily to the lack of observations, mentioning that the input training dataset was small. Alimissis \textit{et al.} \cite{alimissis2018spatial} developed an artificial neural network model and a multiple linear regression model to predict pollutant concentrations in the Attica region in Greece. The selected pollutants included nitrogen dioxide, sulphur dioxide, ozone, carbon monoxide, and nitrogen oxide. The models were evaluated based on cross-validation by leaving one station out of the training dataset, then predicting pollutant concentrations for that station and calculating certain measures, such as root mean squared error, mean absolute error, and coefficient of determination. The authors verified that for most stations, the artificial neural network model outperformed the multiple linear regression model, which is due to the linear regression not being able to model complex relationships within the data. The authors emphasized that the neural network model predictive ability is highly related to the structure of the network and the fine tuning of its parameters. They also acknowledged that the artificial neural network requires extensive input training data, indicating this as one of its biggest disadvantages.
        
        The chosen model to be applied as a first attempt to infer pollutant concentrations in locations where no ground measurements exist and fill in maps is the random forest model. This model can identify and model linear and non-linear relationships within the data, provide accurate predictions, provide support for handling overfitting, and is simpler to implement \cite{bonaccorso2017machine}. Moreover, the random forest model also performs implicit feature selection and provides the importance for each feature in the model, which helps identify noise variables and give some insight about some of the more important variables \cite{bonaccorso2017machine}.
        
        The random forest model has been previously successfully applied in the literature to determine pollutant concentrations. Jie Chen \textit{et al.} \cite{chen2019comparison} developed and compared 16 models, using linear regression, regularization, and machine learning algorithms, to predict annual average PM2.5 and nitrogen dioxide concentrations in Europe. The developed models used as predictors satellite measurements, more specifically, aerosol optical depth and tropospheric nitrogen dioxide columns, chemical transport model results, and land use classification variables, and were evaluated by executing 5-fold cross-validation and external validation using two different ground measurement datasets. The authors verified that for the PM2.5 models, the random forest model and other two similar models performed slightly better than the rest. The authors also acknowledged the ability of machine learning algorithms to model complex spatiotemporal variations within the data. Gongbo Chen \textit{et al.} \cite{chen2018machine} developed a random forest model and two other traditional regression models to estimate daily ground-level concentrations of PM2.5 in China for the years 2005 to 2016. The developed models used as predictors aerosol optical depth retrieved by satellites, meteorological variables, land cover data, and other temporal and spatial variables. The models were evaluated by performing 10-fold cross-validation, each fold using 10\% of the total number of stations as test set (randomly selected) and the rest as training set, utilizing PM2.5 ground measurements from 2014-2016. The authors concluded that the random forest model performed much better than the other two regression models, showing considerably higher predictive ability. According to the authors, despite achieving similar results as other machine learning algorithms, the random forest model distinguishes itself from them due to being simpler to implement and for its user-friendliness. They referred that the user friendliness results from the model simplifying the process of defining complex relationships between predictors and the use of variable importance measures to help the user identify different variables. Rochelle Schneider \textit{et al.} \cite{schneider2020satellite} applied the random forest model to various stages of their study. The main objective of the work was to create a spatio-temporal model capable of inferring daily PM2.5 concentrations across Great Britain, which was achieved in four different stages. In stage 1, a random forest model was developed to predict PM2.5 concentrations from PM10 ground-level measurements, in an attempt to increase the available data for the study. In stage 2, a random forest model was developed to determine aerosol optical depth measured by satellites from reanalysis model results to fill in gaps resulting from retrieval errors and/or missing data from cloud cover problem. In Stage 3, the output from stage 1 and stage 2 is incorporated with spatial and temporal predictors, such as meteorological data, land cover classification, population density, NDVI, variables derived from PM2.5 concentrations, and others, to create a model that predicts PM2.5 using the random forest algorithm. Finally, in stage 4, the developed model is used to determine PM2.5 concentrations in a 1km grid across Great Britain. The models were evaluated based on 10-fold cross-validation, where each fold used 10\% of the stations as test data and the rest as training data. All stages presented good results, with low RMSE and high $R^2$. This study presents some limitations, since it heavily relies on determining PM2.5 concentrations from PM10 measurements and using them as predictors in the final model, and the multi-stage implementation hindering the accuracy of the results of the final model, not allowing a correct quantification of these.

\section{Data sources}

    \subsection{European Air Quality Portal}
    
        \hspace{\parindent}
        Pollutant concentration data was collected from the European Air Quality Portal (EAQP) \cite{EAQP_webpage}. The portal is managed and maintained by the European Environment Agency, and provides pollutant concentration data, from 2000 up to 2022, reported by the EU member states. A script was developed to automatically download and process the files from the EAQP website.

    \subsection{Sentinel-5P}
    
        \hspace{\parindent}
        Sentinel-5P \cite{esa_sentinel5p} is the first satellite of Copernicus, a programme developed by the European Union that aims to collect information about Earth, developed exclusively for observing and monitoring the atmosphere.
        
        Sentinel-5P data is provided at two levels depending on the processing it's been subject to. Level 1-B products consist of geolocated Earth radiances in all spectral bands. Level 2 products consist of geolocated total/tropospheric columns of various pollutants. In this work, the following level 2 products were utilized: Ozone total column (L2\_\_O3\_\_\_\_), Nitrogen Dioxide total and tropospheric columns (L2\_\_NO2\_\_\_), Sulfur Dioxide total column (L2\_\_SO2\_\_\_), and UV Aerosol Index (L2\_\_AER\_AI).
        
        Each Sentinel-5P level 2 product file is in NetCDF format and contains measurements from a single orbit, with resolution 7.2×3.6 km$^2$ before August 6th, 2019, and 5.6×3.6 km$^2$ after this date.
        
        Data download and initial processing was done using the python library S5P-Tools \cite{s5p_tools}. The data is downloaded based on various parameters defined by the user, such as, Sentinel-5P level 2 product, time interval, and region of interest. After the download is completed, the data is processed, converting level 2 products to level 3, and a NetCDF file with the results is created. The output NetCDF file is in a regular latitude/longitude grid with the specified resolution. The selected resolution was 0.03\textdegree x 0.03\textdegree, aiming at creating a regular grid with resolution similar to the highest resolution in the original Sentinel-5P files.
        
        The selected quality value for filtering the pixels was 0.75, which is the recommended value to use, as it doesn't include cloud-covered scenes, scenes covered by snow or ice, problematic retrievals or errors \cite{sentinel5p_pum_no2}. An exception is the UV Aerosol Index (L2\_\_AER\_AI) product. In this case, the selected quality value for filtering the pixels was 0.8, which is the minimum recommended value.
        
        Conversion from level 2 to level 3 products, included in the S5P-Tools script, is achieved with the HARP toolkit \cite{atmospheric_toolbox_harp}. This toolkit is available in various different programming languages and allows to read, process and inter-compare different data sets, such as, remote sensing data, model data, observation data, etc., by creating output files with the same spatial/temporal grid and data structure \cite{atmospheric_toolbox_harp}. 
    
    \subsection{ERA5}
    
        \hspace{\parindent}
        ERA5 is a reanalysis dataset, as it combines model data with observations to produce hourly estimates of various atmospheric and land variables, such as, temperature, humidity, pressure, etc. ERA5 is created by the European Centre for Medium-Range Weather Forecasts and made available in the Climate Data Store \cite{climate_data_store}. ERA5 data is available in numerous different sub sets, including hourly estimates and monthly averages, in both single levels (surface quantities) and different pressure levels (upper air fields). In this work, the "ERA5 hourly data on single levels from 1979 to present" dataset was used \cite{era5_hourly_single_levels}, which contains hourly estimates from 1979 to present in a regular latitude-longitude grid, with 0.25 degrees resolution and global coverage.
        
        A script was developed to automatically download data from the "ERA5 hourly data on single levels from 1979 to present" dataset. The selected variables are the following: 2m dewpoint temperature, 2m temperature, 10m u-component of wind, 10m v-component of wind, Surface solar radiation downwards, Evaporation, Total precipitation, Boundary layer height and Surface pressure.
    
    \subsection{Corine Land Cover}
        
        \hspace{\parindent}
        Corine Land Cover is a land use dataset made available in the Copernicus Land Monitoring Service, that is produced by the European Environment Agency with the purpose of standardizing land data collection in Europe \cite{corine_land_cover_webpage}. The data is created by visually analysing high resolution satellite imagery and is updated every six years, the most recent update being from 2018 \cite{corine_land_cover_webpage}. 

        The dataset provides land classification for all of Europe based on 44 different classes from five main groups: Artificial surfaces, Agriculture, Forests and seminatural areas, Wetlands, and Water \cite{corine_land_cover_webpage}.
        
        The Corine Land Cover dataset was manually downloaded from the Copernicus Land Monitoring Service website \cite{corine_land_cover_webpage}. The downloaded file is a raster in GeoTiff format with 100 metres resolution, representing all of Europe. 

        After downloading the file, a script was developed to obtain the desired classes data for each air quality station considered. The selected classes were the following: 'Continuous urban fabric', 'Discontinuous urban fabric', 'Industrial or commercial units', 'Road and rail networks and associated land', 'Port areas', 'Airports', and 'Broad-leaved forest'.

\section{Random Forest Model}
    \subsection{Model implementation}
        
        \hspace{\parindent}
        A random forest model was developed for every considered pollutant to infer pollutant concentrations in locations where no observations exist in the Iberian Peninsula. 

        The random forest model was implemented using the scikit learn package, a machine learning python library \cite{scikit_learn}. Firstly, the model is defined based on the "RandomForestRegressor" ensemble class of the library, which is initialized with the selected hyperparameters, these being: the maximum number of features utilized for building each tree and the number of estimators in the forest. The hyperparameters were optimized and the optimization process is discussed in section \ref{section:parameter_optimization}. A random state is also defined to be able to acquire consistent results. Afterwards, the model is trained using the "fit" method of the library and training data in tabular format.  

        For each pollutant and year, a dataset was produced to train and test the developed random forest models. The dataset is in csv format and each row in the file contains the feature values for a certain station at the time of the satellite passage. The selected features include spatial, temporal, meteorological and land classification variables to take into account the spatial and temporal variation of pollutant concentrations, as well as the creation, transport and deposition of pollutants.
        
        The variables included in the model were the following:
        \begin{itemize}
          \item Day of the week, ranging from 1 (Monday) to 7 (Sunday);
          \item Day of the year, ranging from 1 to 365;
          \item Satellite passage hour, month, and year;
          \item Station type, with the values 1 (industrial), 2 (traffic) and 3 (background);
          \item Station latitude, longitude, and altitude;
          \item Satellite measurement - "Tropospheric NO2 column number density" for $NO_2$ model, "O3 column number density" for $O_3$ model, "SO2 column number density" for $SO_2$ model, and "Absorbing aerosol index 340/380nm" for $PM10$ and $PM2.5$ models;
          \item Station observation - Ground measurement from air quality station for each specified pollutant ($NO_2$, $O_3$, $SO_2$, $PM10$, $PM2.5$) at the time of passage of the satellite, determined using linear interpolation;
          \item Corine Land Cover variables - area occupied by different relevant land-uses, including Continuous urban fabric, Discontinuous urban fabric, Industrial or commercial units, Road and rail networks and associated land, Port areas, Airports, Broad-leaved forest. For each variable and satellite grid pixel a percentage was determined;
          \item Meteorological variables - Wind speed, Wind direction, Dewpoint temperature, Evaporation, Temperature, Total precipitation, Surface Pressure, Boundary Layer Height, Surface Solar Radiation Downwards at the time of passage of the satellite.
        \end{itemize}
        
    \subsection{Parameter optimization}
    \label{section:parameter_optimization}
    
        \hspace{\parindent}
        Two parameters of paramount importance in the creation of a random forest model are "n\_estimators", which controls the number of trees in the forest, and "max\_features", which defines the minimum number of features to consider when calculating the best split of a node. 

        The "max\_features" parameter can have the following values: 'auto', which considers all available features; 'sqrt', which uses only the square root of the total number of features; and 'log2', which uses only the logarithm of the total number of features.
        
        The parameters were optimized by determining the execution time and the mean squared error in a 3-fold cross-validation for various numbers of trees in the forest, from 50 until 500, with increments of 50, and for each of the identified values of the "max\_features" parameter. The mean squared error is calculated for each fold and then an average is determined. The measured time corresponds to the number of seconds it took to execute the 3-fold cross-validation.
        
        The selected values for the parameters where the ones that resulted in a lower error, taking into account the execution time. The error decreases as the number of estimators in the model increases, however, training and testing time also increases, where a minimal decrease in error value doesn't compensate for the big increase in execution time.   
        
\begin{figure*}[h]

    \begin{minipage}{0.5\linewidth}
        %        \begin{figure}[h]
         \centering
         \includegraphics[width=\columnwidth]{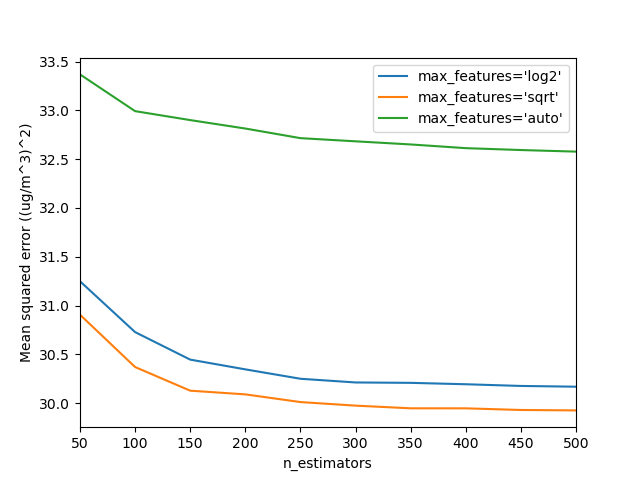}
         \caption{Mean squared error graph for each "max\_features" value and different number of estimators, for the nitrogen dioxide 2019 dataset.}
         \label{fig:Parameter_optimization_NO2_mse}
%        \end{figure}
   \end{minipage} \begin{minipage}{0.5\linewidth}
%        \begin{figure}[h]
            \centering
            \includegraphics[width=\textwidth]{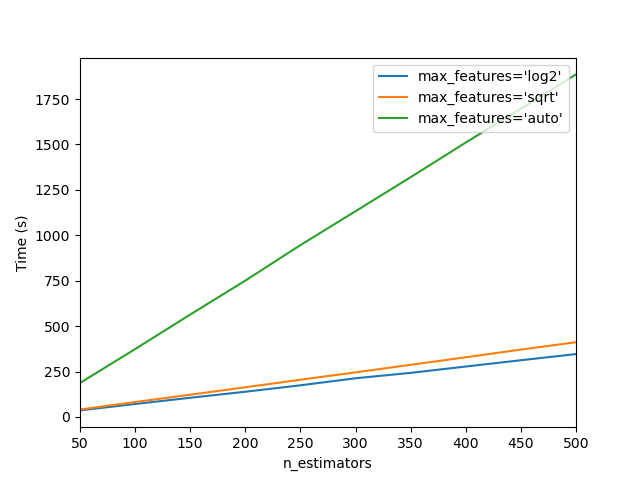}
            \caption{Execution time graph for each "max\_features" value and different number of estimators, for the nitrogen dioxide 2019 dataset.}
            \label{fig:Parameter_optimization_NO2_time}
%        \end{figure}
        \end{minipage}
\end{figure*}

        Model parameters were determined utilizing the produced 2019 datasets for each pollutant. With regards to the nitrogen dioxide 2019 dataset, the obtained graphs for the mean squared error and execution time are presented in Figure \ref{fig:Parameter_optimization_NO2_mse} and Figure \ref{fig:Parameter_optimization_NO2_time}, respectively. In this case, the chosen parameters were "n\_estimators" = 300 and  "max\_features" = 'sqrt'. Similar results were obtained for the other pollutants, where the chosen parameters were also "n\_estimators" = 300 and  "max\_features" = 'sqrt', with the exception of ozone, where "max\_features" = 'auto' produced better results.

\section{Evaluation}

    \subsection{Feature Importance}
    
        \hspace{\parindent}
        
        For each considered pollutant, a random forest model was created using the hyperparameters defined in section \ref{section:parameter_optimization} and the produced 2019 dataset for the Iberian Peninsula, where 80\% of the dataset was used as training data and 20\% as test data. After training the model, Gini importance and Permutation importance measures were obtained. Gini importance measurements are directly extracted after model training and permutation importance measurements are determined with the randomly selected test data. 
        
        Although Gini importance and Permutation importance values are not directly comparable, the observed positioning of the features is generally similar with slight differences.
        
        The most important features for each model vary according to the different datasets. Nonetheless, some spatial features, such as longitude, latitude, and land cover, and temporal features, such as day of year, are consistently considered some of the most important, reflecting the existent spatial and temporal variation of pollutant concentrations. It is also worth noting the importance of the Era5 meteorological variable "Boundary Layer Height" that is considered very important in most models. The boundary layer is the lowest part of the atmosphere and its height is directly influenced by all chemical and physical processes that occur on Earth's surface, where at lower heights, higher concentrations of pollutants may develop \cite{era5_data_documentation}. With regards to the $O_3$ model, the "Surface Solar Radiation Downwards" and "Temperature" were considered as the most important features for the model, which is in accordance with the formation of ground level ozone. This pollutant results from photo chemical reactions between two primary pollutants, and, as a result, higher radiation hitting the Earth's surface and warmer temperatures create more favourable conditions for the formation of ground level ozone \cite{monks2015tropospheric}. 

        The importance of the integration of remote sensing variables in the models varied widely depending on the model. With regards to the $NO_2$ model, the "Tropospheric NO2 column number density" variable was identified as the most important feature by both importance measurement methods, being considered as a strong predictor of surface $NO_2$ concentration. The $NO_2$ pollutant is the only one (of the considered pollutants in this study) that Sentinel-5P provides tropospheric measurements for, while the others only have total column measurements. The remote sensing variable "O3 column number density" was also considered somewhat important for the $O_3$ model, although not as much as the previous one. These results contrast with those obtained for the $SO_2$, $PM10$, and $PM2.5$ models. The remote sensing variable "Absorbing aerosol index 340/380nm" was considered one of the least important in Permutation importance measurements for both $PM10$ and $PM2.5$ models. Usually, aerosol optical depth (AOD) measurements retrieved by satellites are included in models that predict particulate matter concentrations, however, Sentinel-5P doesn't provide any AOD measurements and the aerosol index was used instead. With regards to the $SO_2$ model, the "SO2 column number density" variable was considered as the least important by the Permutation importance method, exhibiting negative importance values. This indicates that by randomly shuffling this variable and eventually by random chance the model presented better performance than before.

    \subsection{Description of model evaluation methods}
    
        \hspace{\parindent} 
        
        The developed random forest models for each pollutant were evaluated temporally and spatially using three different methods.
        
        Method A consists of creating a random forest model, for every considered pollutant, and evaluating it by applying 10-fold cross-validation and calculating error and bias measures.
        
        Method B was created to temporally evaluate the developed random forest models, in order to understand if the models trained with data from one year produce quality results when used to predict pollutant concentrations for another year. In this case, the produced 2018 and 2019 datasets of the considered pollutant were used to test and train the models, respectively. Data from 2018 is used as test data instead of data from 2020 due to the Covid-19 pandemic lockdown severely altering pollutant emissions. Since the main objective of this work was to assess the use of Sentinel-5P products to estimate pollutant concentration, only normal conditions were considered to exclude the effects of the Covid pandemic in model performance.
        
        Method C spatially evaluates the developed models by applying 10-fold cross-validation, modified to use part of the stations as testing data and the rest as training data. In this case, each fold uses 10\% of the total number of stations as a testing set and the rest as training. The determination of the stations to be included in the test set in each fold is accomplished with the intent of being as spatially representative of the study area as possible. The produced 2019 datasets of the considered pollutant are used to train and test the models, and some error and bias measures are obtained. This method is considered to better evaluate the error obtained when estimating pollutant concentrations in areas without observations.
        
    \subsection{Model evaluation results}
    
        \hspace{\parindent}     
        
        All methods were evaluated by calculating error and bias measures. These measures include the coefficient of determination ($R^2$), root mean squared error (RMSE), and bias, whose formulas are presented in Equations \ref{eq:r2}, \ref{eq:rmse}, and \ref{eq:bias}, respectively.

        \begin{equation}
        R^2 = 1 - \dfrac{\sum_{i=1}^{N} (o_{i} - p_{i})^2}{\sum_{i=1}^{N} (o_{i} - \overline{o})^2},
        \label{eq:r2}
        \end{equation}
        
        \begin{equation}
        RMSE = \sqrt{\dfrac{\sum_{i=0}^{N-1} (o_{i} - p_{i})^2}{N}},
        \label{eq:rmse}
        \end{equation}
        
        \begin{equation}
        Bias = \dfrac{\sum_{i=1}^{N} (p_{i} - o_{i})}{N},
        \label{eq:bias}
        \end{equation}
        
        where $o_i$ is the value of observation i, $p_i$ is the value of prediction i, $N$ is the number of samples in the test set, and $ \overline{o}$ is the mean observation value.
        
        The obtained results for method A, method B and method C for all pollutants are presented in Table \ref{table:method_A_results}, Table \ref{table:method_B_results} and Table \ref{table:method_C_results}, respectively.

        \begin{table}[!ht]
        \centering
        \caption{ Method A 10-fold cross-validation results for all developed models (one for each considered pollutant).}
        \label{table:method_A_results}
        \begin{tabular}{ |c||c|c|c| }
         \hline
         Pollutant & Mean $R^2$ & Mean RMSE & Mean Bias \\
         \hline
         $NO_2$ & 0.8181 & 5.2367 & 0.1140 \\
         \hline
         $O_3$ & 0.8552 & 9.5791 & -0.045 \\
         \hline
         $SO_2$ & 0.5248 & 4.3873 & 0.0737 \\
         \hline
         $PM10$ & 0.6295 & 8.1052 & 0.1619 \\
         \hline
         $PM2.5$ & 0.6328 & 4.2875 & 0.0959 \\
         \hline
        \end{tabular}
        \end{table}

        \begin{table}[!ht]
        \centering
        \caption{ Method B results for all developed models (one for each considered pollutant).}
        \label{table:method_B_results}
        \begin{tabular}{ |c||c|c|c| }
         \hline
         Pollutant & $R^2$ & RMSE & Bias \\
         \hline
         $NO_2$ & 0.6786 & 7.0288 & -0.7600 \\
         \hline
         $O_3$ & 0.6759 & 15.7464 & -2.3700 \\
         \hline
         $SO_2$ & 0.3040 & 4.8222 & -0.0820 \\
         \hline
         $PM10$ & 0.3267 & 11.6018 & -0.5369 \\
         \hline
         $PM2.5$ & 0.3417 & 5.9500 & -0.6522 \\
         \hline
        \end{tabular}
        \end{table}

        \begin{table}[!ht]
        \centering
        \caption{ Method C 10-fold cross-validation results for all developed models (one for each considered pollutant).}
        \label{table:method_C_results}
        \begin{tabular}{ |c||c|c|c| }
         \hline
         Pollutant & Mean $R^2$ & Mean RMSE & Mean Bias \\
         \hline
         $NO_2$ & 0.5524 & 7.8176 & 0.4567 \\
         \hline
         $O_3$ & 0.7462 & 12.5934 & -0.3342 \\
         \hline
         $SO_2$ & -0.0231 & 5.9541 & 0.3360 \\
         \hline
         $PM10$ & 0.3722 & 10.4737 & 0.4257 \\
         \hline
         $PM2.5$ & 0.3303 & 5.6711 & 0.3778 \\
         \hline
        \end{tabular}
        \end{table}
        
        %% Analise de resultados
        After analysing Tables \ref{table:method_A_results}-\ref{table:method_C_results}, it can be verified that method A overall exhibits high values of $R^2$, possibly indicating really good models. Nonetheless, method A is not the best for evaluating the developed models. Although pollutant concentration observations present a temporal dependency, varying depending on the time of year, consecutive days tend to have similar measurements. This fact directly affects the results of method A, which will be better than other methods, since the testing set contains observations that are very similar to those in the training set.
        
        Method B presents good values of the coefficient of determination for the $NO_2$ and $O_3$ models, acceptable error measurements and a slight underestimation of the values for the $NO_2$ model and higher underestimation for the $O_3$ model. For the other pollutants the models performed rather poorly. However, method's B results are also not very relevant since the main objective of the creation of the models was to infer pollutant concentrations in locations where no observations exist. 
        
        %% Tentativa de perceber se devia ter modelos anuais ou não?
        
        Method C constitutes the best method for evaluating the developed models, since it directly provides error and bias measures in locations not included in model training, and, as a result, in locations where no observations exist, which is the primary objective of this work. With regards to the station cross-validation coefficient of determination, the $NO_2$ and $O_3$ models presented good values, especially the $O_3$ model. However, the $SO_2$, $PM10$, and $PM2.5$ models performed very poorly in this regard, especially the $SO_2$ that presented a negative value of the coefficient of determination, which indicates that generally, a model that always predicts the mean value of the observations performed better than the developed model. All models slightly overestimated the surface concentrations, indicated by the positive values of the station cross-validation mean bias, except the $O_3$ model where the surface concentrations were generally slightly underestimated. With regards to the RMSE, all models presented acceptable cross-validation errors, except the $O_3$ and $PM10$ models where the mean value was a little higher. 
        
        In all method's results, the coefficient of determination is lower for the $SO_2$, $PM10$, and $PM2.5$ models. This is in accordance with what was observed when analysing model variable importance, since these models don't present any consistently strong predictors, and so, the variation of the dependent variable is not explained by the independent variables, and the coefficient of determination is lower.

    \subsection{Temporal and spatial analysis of model predictions}
    
        \hspace{\parindent}
    
        The developed models that presented satisfying results, such as those built for the $NO_2$ and the $O_3$, were utilized to predict pollutant concentrations for 2019 in the Iberian Peninsula in locations where no observation data exists. To achieve this, a grid was defined and new datasets were created that include desired feature values for all cells in the grid. The chosen grid was the regular grid produced by the HARP toolkit based on an area of interest defined in a geojson file, which includes approximately 75000 pixels with a resolution of 0.03\textdegree x 0.03\textdegree. 
        
        Model predictions were obtained for all satellite passages for each pixel in the grid. The results for each cell were averaged to create a map with an annual mean concentration. Model predictions were also utilized to produce box plots containing the temporal variation of pollutant concentrations.
        
\begin{figure}[!ht]
        \centering
        \includegraphics[width=0.4\textwidth]{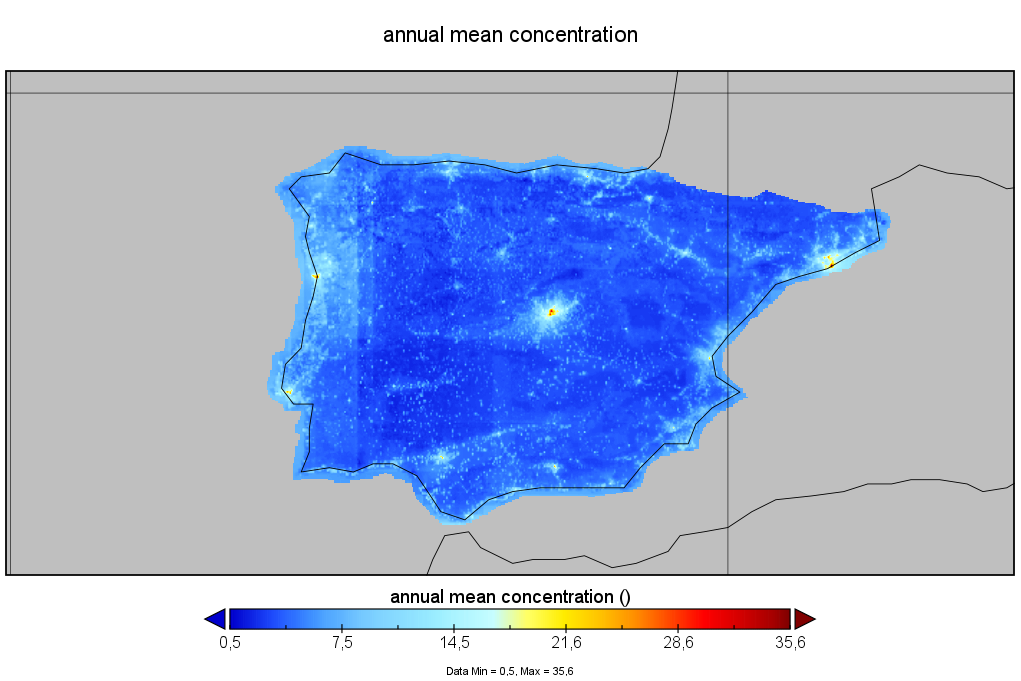}
        \caption{Annual mean concentration of nitrogen dioxide (in $\mu g/m^3$) in 2019 in the Iberian Peninsula, calculated from model predictions.}
        \label{fig:NO2_2019_annual_mean_concentration}
        \end{figure}

\begin{figure}[!ht]
        \centering
        \includegraphics[width=0.4\textwidth]{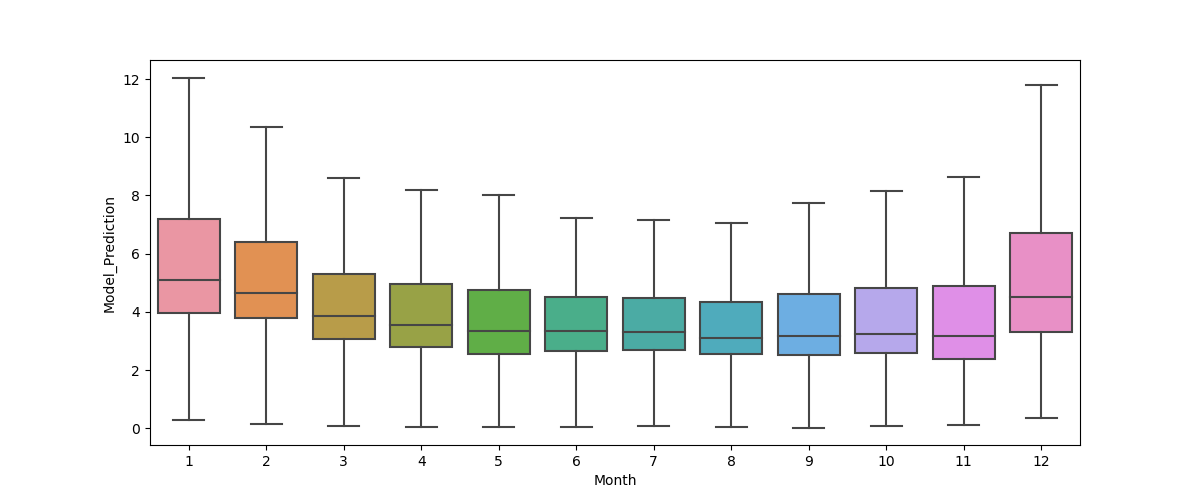}
        \caption{Box plot with temporal variation of nitrogen dioxide concentration (in $\mu g/m^3$) in 2019 in the Iberian Peninsula, obtained from model predictions. Each number corresponds to the respective month number in the year, that is January is 1 and December is 12.}
        \label{fig:NO2_2019_boxplot}
        \end{figure}

        The 2019 annual mean nitrogen dioxide concentration map obtained from model predictions is presented in Figure \ref{fig:NO2_2019_annual_mean_concentration}. The box plot containing the temporal variation of nitrogen dioxide model predictions is presented in Figure \ref{fig:NO2_2019_boxplot}. 

        The main source of nitrogen dioxide is fuel combustion and the main sector that contributes to these emissions is the road transport sector. This can be observed in the annual mean concentration map presented in Figure \ref{fig:NO2_2019_annual_mean_concentration}, in which nitrogen dioxide concentration is much higher in cities, cities outskirts and high density roads, where road traffic is much more intense. Other nitrogen dioxide emissions sources include energy supply/generation and heating, which explains larger values in Figure \ref{fig:NO2_2019_boxplot}, since the temporal variation presented in the box plot indicates higher concentrations of nitrogen dioxide in the winter months (December, month 12; January, month 1; and February, month 2) probably related with heating.
        
        The 2019 annual mean ozone concentration map obtained from model predictions is presented in Figure \ref{fig:NO2_2019_annual_mean_concentration}. The box plot containing the temporal variation of ozone model predictions is presented in Figure \ref{fig:NO2_2019_boxplot}. 

        \begin{figure}[!ht]
        \centering
        \includegraphics[width=0.4\textwidth]{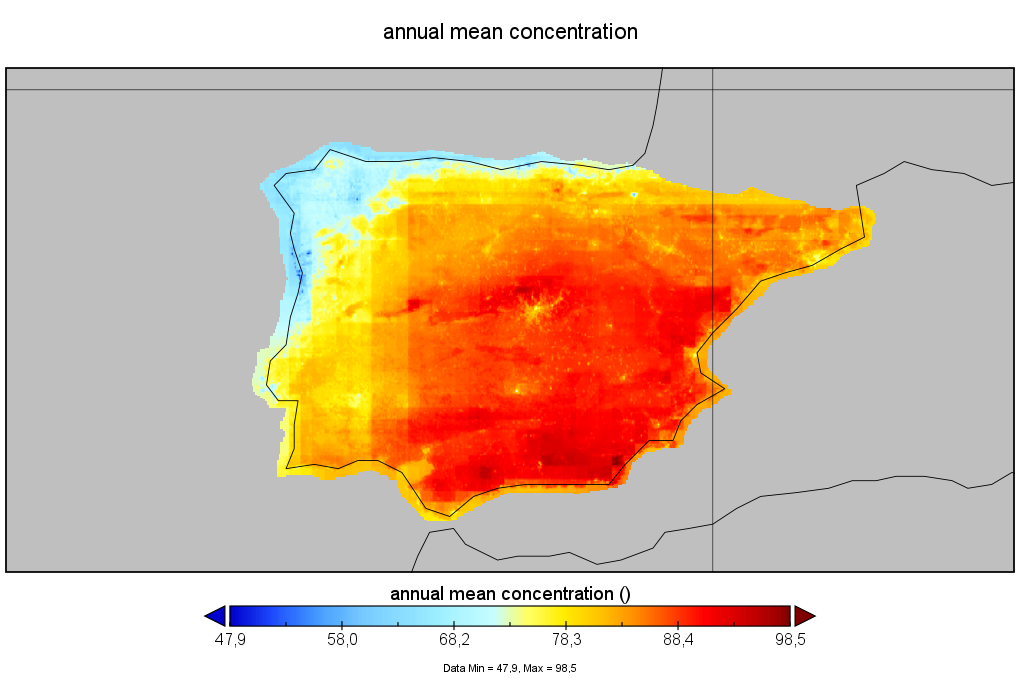}
        \caption{Annual mean concentration of ozone (in $\mu g/m^3$) in 2019 in the Iberian Peninsula, calculated from model predictions.}
        \label{fig:O3_2019_annual_mean_concentration}
        \end{figure}
        
        \begin{figure}[!ht]
        \centering
        \includegraphics[width=0.4\textwidth]{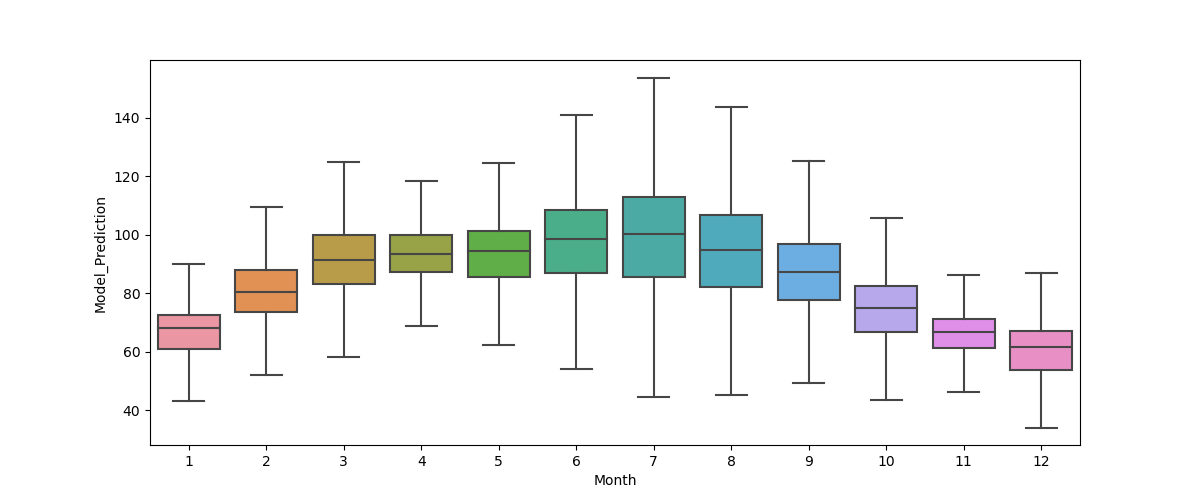}
        \caption{Box plot with temporal variation of ozone concentration (in $\mu g/m^3$) in 2019 in the Iberian Peninsula, obtained from model predictions. Each number corresponds to the respective month number in the year, that is January is 1 and December is 12.}
        \label{fig:O3_2019_boxplot}
        \end{figure}
        
        Ground level ozone is formed due to photo chemical reactions, that is, chemical reactions that occur as a result of light absorption, between volatile organic compounds and nitrogen oxides, both primary pollutants mainly emitted into the atmosphere by chemical plants (e.g., power plants) and combustion processes (e.g., vehicle fuel combustion). Nonetheless, ground level ozone reacts with nitrogen oxide ($NO$) to form nitrogen dioxide ($NO_2$) and oxygen ($O_2$), which leads to ozone degradation in cities, since nitrogen oxide is largely emitted from fuel combustion. Therefore, larger cities will present lower ozone concentrations than rural areas, in which there are much less emissions of $NO$ (causing less degradation of ozone). This can be observed in the annual mean concentration map presented in Figure \ref{fig:O3_2019_annual_mean_concentration}, showing a south-increasing trend, following trends in temperature and also locally showing lower mean values overlapping larger cities. Ground level ozone concentration is higher during the summer months due to more radiation hitting the Earth's surface, warmer temperature and longer daylight periods, creating more favorable conditions for the occurrence of the photo chemical reactions that form ground level ozone. This can also be observed in Figure \ref{fig:O3_2019_boxplot}, in which the box plot with the temporal variation of ozone concentration indicates higher concentrations in June (month 6), July (month 7), and August (month 8).

\section{Conclusion}

        Air pollution is a global problem that severely affects not only human health, but also the environment. The existence of spatial and temporal data regarding the concentrations of pollutants is crucial for performing air pollution studies and monitor emissions. 

        Ground monitoring stations provide reliable data on the concentration of pollutants with great temporal resolution. Nonetheless, the number of stations is very limited and are usually built in more densely populated areas, leaving a large portion of the territory without any pollutant concentration measurements \cite{european2011application}. 
        
        Remote sensing measurements can be used to complement observations, since they provide global coverage with good spatial resolution. Remote sensing provides data on the total column of a substance (integrated concentration from Earth's surface to the top of the atmosphere \cite{veefkind2007applicability}).
        
        In this work, a random forest model was developed to infer pollutant concentrations for 2019 in the Iberian Peninsula in locations where no observations exist. A model was developed for each one of five selected pollutants: $NO_2$, $O_3$ $SO_2$, $PM10$, and $PM2.5$. To account for the temporal and spatial variation of pollutant concentrations, as well as, the direct effect of certain variables on the creation and transport of pollutants, the developed models include satellite measurements, meteorological variables (wind speed, temperature, humidity, etc.), land use classification (mostly identifying emission sources), spatial features (latitude, longitude, altitude), and temporal features (day of the year, week day, etc.).  
        
        The relevance of the integration of remote sensing variables in the models varied significantly depending on the model, with some variables having much more influence than others. The "Tropospheric NO2 column number density" variable was identified as the most important feature in the $NO_2$ model, being considered as a strong predictor of surface $NO_2$ concentrations. With regards to the $O_3$ model, the meteorological variables ”Surface Solar Radiation Downwards” and ”Temperature” were considered as the most important features for this model, while the remote sensing variable "O3 column number density" was considered somewhat important but not as much as the previous one. The results were very poor with regards to the $SO_2$, $PM10$, and $PM2.5$ models. The variables "Absorbing aerosol index 340/380nm" and "SO2 column number density" were considered one of the least important in the $PM10/PM2.5$ and $SO_2$ models, respectively.
        
        From all the developed models, the $NO_2$ and $O_3$ were the ones that presented the best results after applying station 10-fold cross-validation. These models were used to predict surface concentrations in 2019 in the Iberian Peninsula. After analysing spatially and temporally the model predictions, many patterns relating to the emission/formation of $NO_2$ and $O_3$ respectively were identified, further solidifying the developed models. The station 10-fold cross-validation results for the $SO_2$, $PM10$, and $PM2.5$ models were very poor, with these being considered unfit for usage.

        Regarding the integration of remote sensing, total column measurements could be replaced by tropospheric measurements when made available by Sentinel-5P or by other satellites, since the main objective is to determine surface pollutant concentrations and tropospheric measurements can provide a better relation to what is happening at the surface level. As observed for the $NO_2$ model, the "tropospheric NO2 column number density" variable was identified as the most important feature in this model. The $NO_2$ pollutant is the only one (of the considered pollutants in this study) that Sentinel-5P provides tropospheric measurements for, while the others only have total column measurements.

        With regard to the $PM10$ and $PM2.5$ models, it is suggested the utilization of aerosol optical depth (AOD) measurements from other satellites. In this work, the Sentinel-5P "Absorbing aerosol index 340/380nm" variable was included in the model since this satellite doesn't provide any AOD measurements. However, the absorbing aerosol index is more fit for identifying the presence of layers of aerosols with significant absorption in the UV spectral range (e.g., desert dust, biomass burning and volcanic ash plumes) \cite{sentinel_aerosol_index}.
        
        With respect to the $O_3$ model, to increase model performance it is suggested the utilization of a meteorological dataset with better spatial resolution. The meteorological variables ”Surface Solar Radiation Downwards” and ”Temperature” were considered as the most important features for this model, which are highly important in the formation of ground level ozone. However model predictions were underestimated due to the relatively low spatial resolution of the Era5 dataset, which presented a pixel size of approximately 26km. Better results can possibly be achieved using a dataset with higher resolution to detect local trends.
        
        More variables, especially spatial variables such as, population density, road density and distance to emission sources, in an attempt to increase model performance. In addition, chemical transport model results could also be included as features in the models, since they provide pollutant concentration measurements with good temporal and spatial resolution.

\bibliographystyle{IEEEtran}
\bibliography{./Bibliography}

\end{document}